\documentclass[runningheads]{llncs}
\usepackage{amsmath} 
\usepackage[utf8]{inputenc}
\usepackage{amssymb}
\usepackage{graphicx}
\usepackage{multirow}
\usepackage[T1]{fontenc}

\usepackage{graphicx,verbatim}
\usepackage{hyperref}
\usepackage{color}

\usepackage{xcolor}
\usepackage{booktabs}

\begin{document}
\title{Two-Stage Decoupling Framework for Variable-Length Glaucoma Prognosis}
%

\author{Yiran Song\inst{1} \and
Yikai Zhang\inst{1} \and
Silvia Orengo-Nania\inst{1} \and
Nian Wang\inst{2} \and
Fenglong Ma\inst{3} \and
Rui Zhang\inst{1} \and
Yifan Peng\inst{4} \and
Mingquan Lin\inst{1}}

\authorrunning{Y. Song et al.} 

\institute{University of Minnesota, Minneapolis MN, USA
\email{\{songyiran,zhan9191,sorengon,zhan1386,lin01231\}@umn.edu} \and
UT Southwestern Medical Center, Dallas TX, USA
\email{nian.wang@UTSouthwestern.edu} \and
The Pennsylvania State University, University Park PA, USA
\email{fenglong@psu.edu} \and
Weill Cornell Medicine, New York NY, USA
\email{yip4002@med.cornell.edu}}  
\maketitle          
\begin{abstract}

Glaucoma is one of the leading causes of irreversible blindness worldwide. Glaucoma prognosis is essential for identifying at-risk patients and enabling timely intervention to prevent blindness. Many existing approaches rely on historical sequential data but are constrained by fixed-length inputs, limiting their flexibility. Additionally, traditional glaucoma prognosis methods often employ end-to-end models, which struggle with the limited size of glaucoma datasets. To address these challenges, we propose a Two-Stage Decoupling Framework (TSDF) for variable-length glaucoma prognosis. In the first stage, we employ a feature representation module that leverages self-supervised learning to aggregate multiple glaucoma datasets for training, disregarding differences in their supervisory information. This approach enables datasets of varying sizes to learn better feature representations. In the second stage, we introduce a temporal aggregation module that incorporates an attention-based mechanism to process sequential inputs of varying lengths, ensuring flexible and efficient utilization of all available data. This design significantly enhances model performance while maintaining a compact parameter size. Extensive experiments on two benchmark glaucoma datasets—the Ocular Hypertension Treatment Study (OHTS) and the Glaucoma Real-world Appraisal Progression Ensemble (GRAPE), which differ significantly in scale and clinical settings, demonstrate the effectiveness and robustness of our approach. Code is available: \href{https://github.com/zongzi13545329/MICCAI.git}{MICCAI LMID Repository}.


\end{abstract}

\section{Introduction}

Glaucoma is one of the leading causes of irreversible blindness worldwide~\cite{bourne2013causes}. Unlike glaucoma detection~\cite{de2023airogs,fan2023one,zhou2023representation,lin2022automated,luo2023harvard}, which evaluates the current state of the disease based on existing data, glaucoma prognosis analyzes historical data to anticipate future disease progression. With the increasing global prevalence of glaucoma, accurate prognostic models are crucial to identify at-risk patients, enabling timely interventions, and advancing personalized treatments to prevent vision loss~\cite{tham2014global,hu2023glim,li2020deepgf}. Several deep learning-based methods have been proposed for the prognosis of glaucoma~\cite{hemelings2020accurate,kamal2022explainable,singh2022novel,singh2022deep,khalil2014review,zhang2023application}. For instance, Lin et al.~\cite{lin2022multi} introduced a multi-scale multi-structure Siamese network (MMSNet) to estimate glaucoma progression using only the first and most recent fundus images. However, relying on only two images limits the ability to fully capture disease evolution over time. Li et al.~\cite{li2020deepgf} constructed a dataset comprising sequential fundus images and developed DeepGF, an LSTM-based model that learns spatial-temporal patterns from the fundus image sequence of a patient. DeepGF predicts the probability of progression of glaucoma in the next step, but does not accurately predict when that progression will occur. To improve temporal modeling, Hu et al.~\cite{hu2023glim} proposed GLIM-Net, a Transformer-based network designed for irregularly sampled fundus image sequences, incorporating two time-related modules to regulate predictions.

\begin{figure}[tb]
\centering
\includegraphics[width=0.8\textwidth]{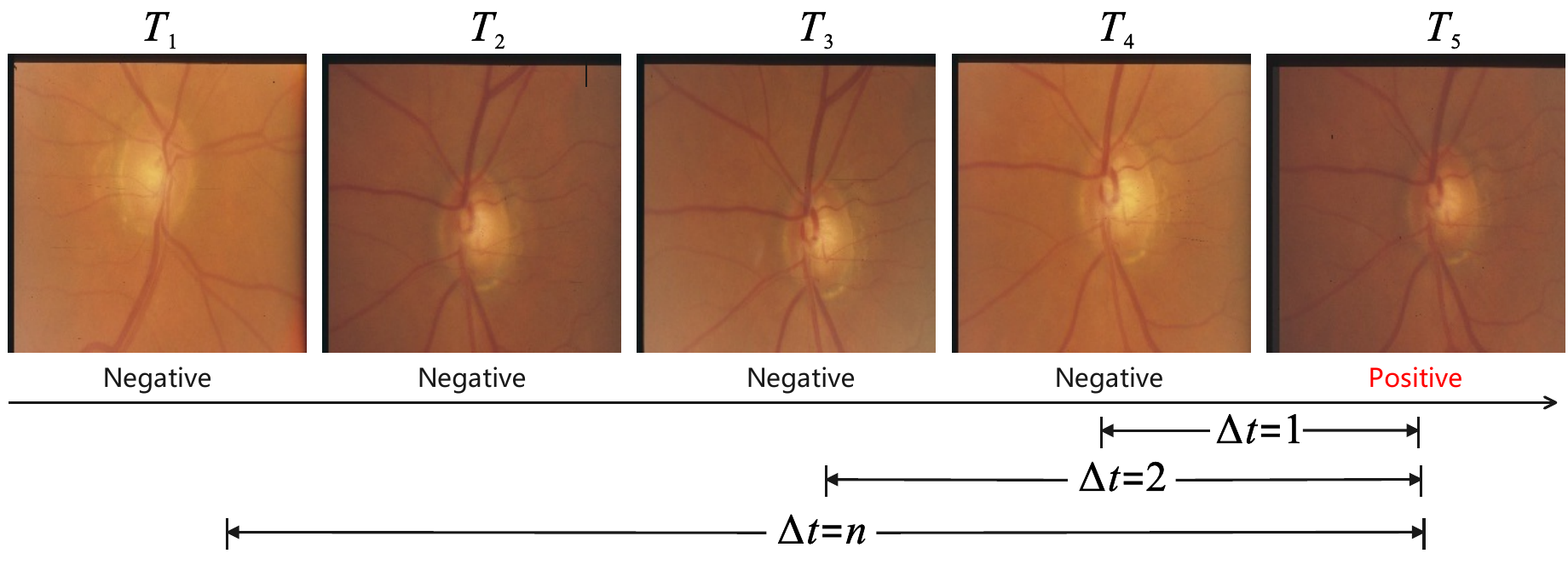}
\caption{Sequential fundus images of a patient of $\Delta t$} \label{T}
\end{figure}

Despite these advancements, two key challenges remain in glaucoma prediction:   \textbf{1. The rigidity of fixed-length time series design significantly affects model flexibility.} Prognosis models with fixed-length designs struggle because glaucoma datasets exhibit a wide range of temporal spans across samples, making them highly sensitive to the chosen sequence length $\Delta t$ (Figure~\ref{T}). A small $\Delta t$ may result in insufficient accuracy while underutilizing longer-sequence data. Conversely, a larger $\Delta t$ may cause some samples to be discarded, while excessive padding can degrade training performance. \textbf{2. The conflict between the large parameter scale of end-to-end models and the limited size of the glaucoma dataset.}  End-to-end models require simultaneous learning of both feature extraction and prediction components. However, glaucoma datasets are typically small, and the need for longitudinal data to train sequence models further reduces the number of available patient-level samples. This constraint makes it difficult to effectively train models with large parameters, increasing the risk of severe overfitting.

\begin{figure}[tb]
\centering
\includegraphics[width=0.9\textwidth]{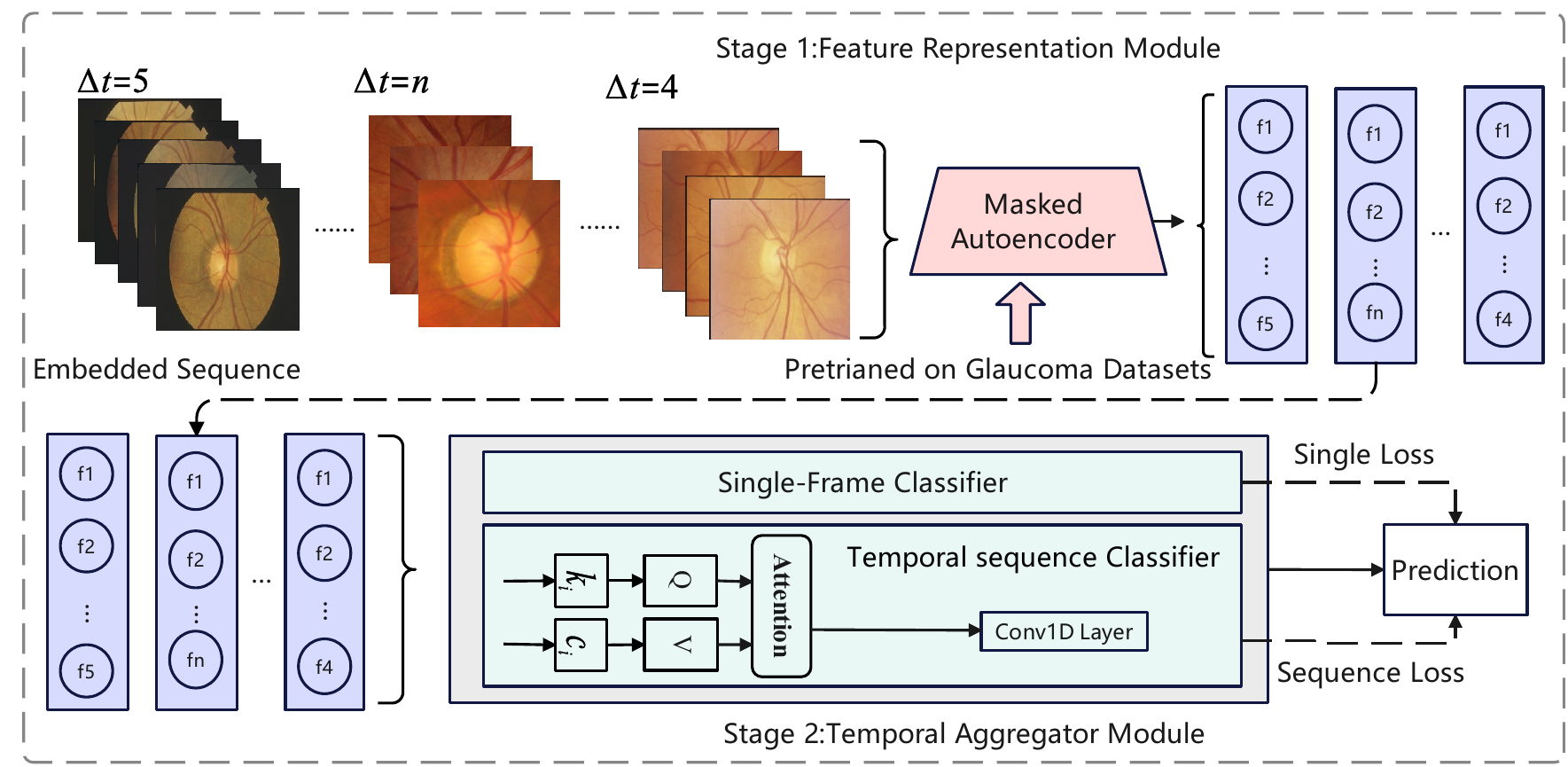}
\caption{The architecture of Our Two-Stage Decoupling Framework (TSDF).} \label{main}
\end{figure}

To address these challenges, we propose a two-stage decoupling framework (TSDF) for glaucoma prognosis (Figure~\ref{main}) that separates feature representation from temporal aggregation. The representation learning module leverages self-supervised learning to fully utilize the available dataset and generate effective representations. Since self-supervised learning does not require labeled data, it enables training on multiple datasets of related diseases, enhancing feature quality for glaucoma prognosis. The temporal aggregator module incorporates an attention-based mechanism, efficiently aggregating variable-length time-series information while reducing model complexity and mitigating overfitting. Its dual-path design captures both single-frame and sequential-frame information. To the best of our knowledge, TSDF is the first framework capable of handling variable-length time-series inputs while simultaneously capturing both single-frame and time-series frame information for glaucoma prognosis. We evaluate TSDF on two publicly available datasets, the Ocular Hypertension Treatment Study (OHTS) ~\cite{gordon2002ohts,holste2024harnessing} and the Glaucoma Real-world Appraisal Progression Ensemble (GRAPE)~\cite{huang2023grape}, demonstrating its effectiveness in handling variable-length sequences and achieving superior prediction performance. In contrast to traditional prediction models, our approach employs a decoupled training strategy that separates feature encoding and diagnosis. This approach enables small datasets to benefit from the extensive knowledge of large datasets during feature encoding, achieving superior patch encoding performance. Additionally, the diagnosis stage requires fewer training parameters, which reduces computational demands and enhances training efficiency, better aligning with practical application requirements.

\section{Method}

\subsection{Self-Supervised Masked Autoencoders for Feature Representation Learning}
We propose leveraging self-supervised learning via a masked autoencoder (MAE) for feature extraction. Specifically, we employ the MAE~\cite{he2022masked} framework, a state-of-the-art self-supervised approach that enables the model to learn robust feature representations by reconstructing masked portions of the input data. The MAE randomly masks a significant fraction of the input patches. The encoder is trained to extract meaningful latent representations, while a lightweight decoder reconstructs the missing content. The training objective is to minimize the reconstruction error, encouraging the encoder to capture high-level semantic features. Once trained, the decoder is discarded, and the encoder is retained to generate feature representations. The datasets used for MAE training consist of patches densely sampled from a single image of temporal data. In each training iteration, a random subset of patches is masked to enhance feature learning and generalization. 

The key advantage of this approach is that it does not require additional supervisory information, relying solely on the inherent structure of the data. This characteristic enables its application to glaucoma datasets with varying levels of supervision, allowing for the integration of multiple datasets for joint training. By aggregating multiple datasets for joint training, the model benefits from increased data diversity and sample size, thereby reducing the likelihood of overfitting typically observed with limited training data.

\subsection{The Dual-path Temporal Aggregator Module}

Unlike previous approaches that separately learn single-frame classifiers or sequence classifiers, TSDF utilizes a dual-path architecture to \textbf{simultaneously optimize both the single-frame classifier and the sequence classifier}. 

Let $T = \{I_0, ..., I_{M-1}\}$ represent a temporal sequence of frames. Given a feature representation function $\phi$, each frame $y_i$ is transformed into an embedding:
\begin{equation}
\mathbf{f}_i = \phi(I_i) \in \mathbb{R}^{D \times 1}, \quad i = 0, ..., M-1.
\end{equation}

\textbf{Single-Frame Classification Path:}
The first path applies a single-frame classifier to each frame embedding:
\begin{equation}
Output_{single}(T) = \mathbf{L}_0 \mathbf{f}_0, ..., \mathbf{L}_{M-1}\mathbf{f}_{M-1}\,
\end{equation}
where the weight matrix $\mathbf{L}_i$ is the instance classifier on each instance embedding. 

\textbf{Sequence Classification Path:} It aggregates frame embeddings into a sequence representation. Each frame embedding $\mathbf{f}_i$ is transformed into two vectors:
\begin{equation}
\mathbf{k}_i = \mathbf{A}_k \mathbf{f}_i, \quad \mathbf{c}_i = \mathbf{A}_c \mathbf{f}_i, \quad i = 0, ..., M-1.
\end{equation}
Here, $\mathbf{k}_i$ serves as the query representation, while $\mathbf{c}_i$ captures the content representation. And we compute the relationship matrix $B$ as follows:
\begin{equation}
B = Q^T V, \quad Q = \begin{bmatrix} \mathbf{k}_0 & \mathbf{k}_1 & ... & \mathbf{k}_{M-1} \end{bmatrix}, \quad V = \begin{bmatrix} \mathbf{c}_0 & \mathbf{c}_1 & ... & \mathbf{c}_{M-1} \end{bmatrix}.
\end{equation}

The resulting sequence representation $\mathbf{B}$ maintains a consistent dimension of $[D \times D]$ \textbf{regardless of sequence length}. Finally, the sequence classification score is computed as:
\begin{equation}
Output_{seq}(T) = \mathbf{V}_2 \text{Conv1D}(B).
\end{equation}
where $\mathbf{V}_2$ is a weight vector applied after convolution.

We utilize the outputs from both the single-frame classifier $Output_{single}(T)$ and the sequence classifier $Output_{seq}(T)$ to simultaneously supervise the model's learning. We used the diagnostic information from a single image as the label for the single-frame classifier and the final prediction result as the label for the sequence classifier. The final loss function is defined as:
\begin{equation}
    \text{loss}_{\text{final}} = \lambda_1 \text{loss}_{\text{single}} + \lambda_2 \text{loss}_{\text{seq}},
\end{equation}
where $\text{loss}_{\text{single}}$ and $\text{loss}_{\text{seq}}$ are the cross-entropy losses from the single-frame classifier and sequence classifier, respectively. We use the coefficients $\lambda_1$ and $\lambda_2$ to control the contribution of these two components to the total loss function.

\section{Experiment}

\subsection{Datasets and Implementation Details}
The OHTS~\cite{gordon2002ohts,holste2024harnessing} dataset represents a landmark multicenter clinical investigation conducted across 22 sites in 16 U.S. states to investigate the progression to primary open-angle glaucoma (POAG). The study enrolled 1,636 participants aged 40-80 years, with intraocular pressure readings of 24-32mmHg in one eye and 21-32mmHg in the contralateral eye. Annual color fundus photography was performed, with POAG status assessments conducted at a centralized Optic Disc Reading Center using a standardized evaluation protocol. The assessment methodology involved independent evaluations by two masked certified readers, with discordant cases adjudicated by a senior masked reader. A quality assurance analysis of 86 eyes demonstrated substantial inter-rater reliability ($\kappa = 0.70$, 95\% CI: 0.55--0.85). Although annual follow-up was targeted, actual visit intervals varied based on participant availability. We utilized 37,399 fundus images, along with corresponding temporal data (in yearly intervals) and POAG diagnostic outcomes for analysis. We excluded terminal visit images for each eye, retaining only the initial image in cases of multiple same-visit acquisitions. The resulting cohort included 30,932 images from 1,597 participants. The dataset maintains a realistic distribution of negative cases, comprising 20\% of the cohort across training, testing, and validation sets, reflecting real-world clinical proportions.

The GRAPE~\cite{huang2023grape} dataset encompasses 1,115 clinical encounters derived from 263 eyes of 144 patients with confirmed glaucoma diagnoses. Longitudinal follow-up included 3-9 visits per eye, with a minimum inter-visit interval of 5 months. The study population has a mean age of 42.49 years, with a balanced distribution across gender and ocular laterality. We perform visual field (VF) progression prediction on the GRAPE dataset. 
Several approaches exist for identifying glaucoma progression in visual fields~\cite{vesti2003comparison,saeedi2019agreement}.
This dataset incorporates three widely adopted automated assessment methods optimized for Octopus perimetry data. Two methods utilize point-wise linear regression (PLR) analysis, where progression is confirmed if either two (PLR2) or three (PLR3) points exhibit significant negative slopes ($P < 0.01$). The third method assesses mean deviation (MD) trends over time, identifying progression when a statistically significant negative MD slope is detected ($P < 0.05$).

\textbf{Data Preprocessing and Fairness Considerations.} We applied unified preprocessing to the dataset to ensure fair comparison. Specifically, for samples that were later converted to positive, we only used their earlier negative samples to ensure that no glaucoma-positive samples were present in the input images. For both baseline methods and our approach, we used identical preprocessing protocols, inputting the temporally last single image, which ensures consistency in the interval between model input and prediction time points, thereby guaranteeing the fairness of results. We selected LSTM and ResNet as baseline methods because current state-of-the-art approaches (e.g., DeepGF\cite{li2020deepgf}, GLIM-Net\cite{hu2023glim}) require additional information inputs (such as polar-transformed fundus images or attention maps) that our model does not need. These input disparities would prevent fair comparison.

TSDF has two core components: the feature representation module and the temporal aggregator module.
\textbf{Feature Representation Module.} We employed the AdamW optimizer ($\beta_1=0.9$, $\beta_2=0.95$) with weight decay of 0.05, except for bias and normalization layers. The base learning rate was set to  $1.5\times10^{-4}$, scaled linearly based on batch size (normalized to 256), with a 40-epoch warmup. Training lasted 400 epochs total. The input images were resized to $224\times224$  pixels with data augmentation including random cropping (scale range 0.2-1.0, bicubic interpolation) and horizontal flipping. Images were normalized using ImageNet statistics (mean=[0.485, 0.456, 0.406], std=[0.229, 0.224, 0.225]). We used ViT-Base~\cite{dosovitskiy2021image} as the backbone with a $16\times16$ patch size and 75\% masking ratio.

\textbf{Temporal Aggregator Module.} The temporal aggregator was implemented in PyTorch with AdamW optimizer ($\beta_1=0.5, \beta_2=0.9$) and trained using cross-entropy loss. The initial learning rate was $10^{-4}$ with cosine annealing scheduling decaying to $5 \times 10^{-5}$. Model weights were orthogonally initialized for all linear and convolutional layers. The feature dimension was set to 768. Training lasted a maximum of 50 epochs with early stopping if no improvement occurred for 10 consecutive epochs. Model selection was based on validation accuracy, with a weight decay of $10^{-3}$ maintained throughout the training process.

We compared our method against two baselines: a ResNet50-based model\cite{he2016deep} and an LSTM-based temporal model\cite{hochreiter1997long}. The ResNet50 model, pre-trained on ImageNet, was fine-tuned for glaucoma prediction using single fundus images. In contrast, the LSTM model required a fixed-length sequence of $\Delta t$ fundus images, using ResNet50 as a feature extractor, followed by bidirectional LSTM layers for temporal modeling. This design caused sample loss when patients had fewer than $\Delta t$ images. To assess the impact of sequence length, we tested LSTM performance with $\Delta t$ ranging from 1 to 4 for OHTS and 1 to 2 for GRAPE, as GRAPE samples generally contain shorter temporal sequences.

In this study, we used 5-fold random cross-validation in all experiments. In each cross-validation, patients were stratified into training (70\%), validation (10\%), and test (20\%) sets at the participant level. Glaucoma prognosis performance was evaluated using accuracy (ACC) and area under the ROC curve (AUC). Statistical significance was assessed using the Mann–Whitney U test, with a $p$-value indicating the probability of observing the difference between models under the null hypothesis. All experiments were conducted on NVIDIA GPUs with CUDA acceleration, implemented in PyTorch with gradient clipping at 5.0 for stable training.

\subsection{Main Results}

\textbf{Quantitative Comparison on OHTS dataset}
As shown in Table~\ref{tab:ohts_results}, ResNet50 achieved the lowest performance metrics (ACC=$0.88$, AUC=$0.78$), primarily due to insufficient information from a single image for effective glaucoma prognosis. For the LSTM temporal model, increasing $\Delta t$ improved both AUC and ACC (ACC increased from $0.86$ to $0.89$, AUC from $0.70$ to $0.83$), while significantly reducing usable sample size (from $3167$ to $2763$). Our approach significantly outperforms the baselines, achieving the highest AUC of $0.928$ and ACC of $0.90$ ($p < 0.01$). This improvement stems from its variable-length design, which utilizes all available samples for decision-making. Additionally, our method employs a decoupling strategy, resulting in a highly efficient model with only 4.1 million parameters—one-sixth the size of ResNet50 (23.5 million parameters) and LSTM (25.9 million parameters). This lightweight design enhances training efficiency and simplifies deployment.

\begin{table}[tb]
   \centering
   \setlength{\tabcolsep}{5pt}
   \caption{The results on the OHTS dataset}
   \begin{tabular}{lccccc}
       \toprule
       & $\Delta t$ & ACC & AUC & PARAMs & Patient Num. \\ 
       \midrule
       Resnet50 & -- & 0.88 & 0.78 & 23,512,130 & 3167 \\ 
       \midrule
       \multirow{4}{*}{LSTM} & 1 & 0.861 & 0.704 & \multirow{4}{*}{25,869,890} & 3167 \\
       & 2 & 0.884 & 0.834 & & 3038 \\
       & 3 & 0.892 & 0.759 & & 2888 \\
       & 4 &  0.898 & 0.863 & & 2763 \\
       \midrule
       TSDF & -- & \textbf{0.907} & \textbf{0.931} & \textbf{4,136,452}& \textbf{3167} \\
       \bottomrule
   \end{tabular}
   \label{tab:ohts_results}
\end{table}

\begin{table}[tb]
\centering
\setlength{\tabcolsep}{5pt}
\caption{The results on the GRAPE dataset.}
\begin{tabular}{lcccccccc}
\toprule
 & $\Delta t$ & \multicolumn{3}{c}{ACC} & \multicolumn{3}{c}{AUC} & \multirow{2}{*}{Patient Num.} \\ \cmidrule(rl){3-5}\cmidrule(rl){6-8}
 &  & PLR2 & PLR3 & MD & PLR2 & PLR3 & MD &  \\ \midrule
Resnet50 & -- & 0.750 & 0.910 & 0.810 & 0.710 & 0.800 & 0.730 & 263 \\ 
\midrule
\multirow{2}{*}{LSTM} & 1 & 0.774 & 0.925 & 0.943 & 0.733 & 0.744 & 0.519 & 263 \\
 & 2 & 0.775 & 0.945 & 0.905 & 0.481 & 0.666 & 0.882 & 196 \\
\midrule
TSDF & -- & \textbf{0.896} & \textbf{0.951} & 0.911 & \textbf{0.866} & \textbf{0.956} & \textbf{0.917} & \textbf{263} \\ \bottomrule
\end{tabular}
\label{tab:grape_results}
\end{table}

\noindent\textbf{Quantitative Comparison on GPAPE dataset}
As shown in Table ~\ref{tab:grape_results}, our method significantly outperforms the baselines, achieving the highest AUCs of 0.896, 0.951, and 0.917 for the PLR2, PLR3, and MD tasks, and the highest ACCs of 0.896 and 0.951 for the PLR2 and PLR3 tasks  ($p < 0.01$). These results demonstrate significant performance improvements compared to the baseline while maintaining a highly efficient parameter configuration. 

OHTS is a large multicenter clinical
trial (1,597 participants, 22 locations), while GRAPE is a real-world dataset
with different diagnostic criteria and a smaller cohort (144 patients). GRAPE
includes three progression labels derived from different visual field assessment
methods (PLR2, PLR3, MD). Our model’s consistent performance across both
datasets demonstrates strong generalizability and robustness.
\subsection{Ablation Study}
\textbf{The ablation study of decoupling design}
To demonstrate the effectiveness of our decoupling design, which separates feature extraction from prognosis, we compared our method with an end-to-end model using ResNet50 as the feature encoder. All settings remained identical except for the feature encoder and training logic (end-to-end vs. decoupling). The end-to-end model achieved 0.891 in accuracy and 0.505 in AUC (OHTS), as well as 0.947 in accuracy and 0.771 in AUC (GRAPE-PLR2), which are notably lower than our method's 0.900 in accuracy and 0.930 in AUC (OHTS) and 0.950 in accuracy and 0.956 in AUC (GRAPE-PLR2). These results confirm that our approach, leveraging self-supervised training and decoupled distribution training, better captures effective information and significantly improves accuracy.

\textbf{The ablation study of Temporal Aggregator Module design}
The temporal aggregator module was trained using both intermediate module outputs and final predictions as loss functions. Balancing their contributions to the overall loss is crucial. Therefore, we investigated the impact of the loss function coefficients $\lambda_1$ and $\lambda_2$, as detailed in Table~\ref{tab:ablation_ohts_grape}. The configuration $\lambda_1=1.5$, $\lambda_2=1$ achieved the best AUC and ACC performance. On the OHTS dataset, it achieved an ACC of 0.891 and an AUC of 0.928, while on the GRAPE dataset, it achieved an ACC of 0.947 and an AUC of 0.956 for the PLR3 task.

\begin{table}[tb]
    \centering
    \setlength{\tabcolsep}{5pt}
    \caption{The ablation study of $\lambda_1$ and $\lambda_2$ in loss function on OHTS and GRAPE datasets}
    \begin{tabular}{cccccc}
        \toprule
        \multirow{2}{*}{$\lambda_1$: $\lambda_2$} & \multicolumn{2}{c}{OHTS} & \multicolumn{2}{c}{GRAPE (PLR3)} & \multirow{2}{*}{PARAMs} \\ 
        \cmidrule(rl){2-3}\cmidrule(rl){4-5}
        & ACC & AUC & ACC & AUC &  \\ 
        \midrule
        0:1 & 0.877 & 0.920 & 0.939 & 0.948 & \multirow{5}{*}{4,136,452} \\
        0.5:1 & 0.890 & 0.913 & 0.947 & 0.953 &\\
        1:0 & 0.579 & 0.585 & 0.425 & 0.426 &  \\
        1:1 & 0.889 & 0.919 & 0.947 & 0.956 &  \\
        1.5:1 & \textbf{0.891} & \textbf{0.928} & \textbf{0.947} & \textbf{0.956} &  \\
        \bottomrule
    \end{tabular}
    \label{tab:ablation_ohts_grape}
\end{table}

Next, we investigated the attention layer designed to balance performance and parameter efficiency. A larger parameter size enhances model robustness but is harder to train on small datasets and increases computational costs. Conversely, a smaller parameter size improves training efficiency but may reduce accuracy on large datasets. As shown in Table~\ref{tab:ablation2_ohts_grape}, we explored three approaches: (1) setting the intermediate layer size to match the input size while maintaining the original attention design,(2) adjusting the intermediate layer to double the input size, and (3) adding a self-attention layer to better capture temporal dependencies. Ultimately, for glaucoma prediction, the design with twice the input size achieved the best performance while maintaining a balanced parameter size.

\begin{table}[tb]
\centering
\setlength{\tabcolsep}{5pt}
\caption{The ablation study of the temporal aggregator module on OHTS and GRAPE datasets.}
\begin{tabular}{lccccc}
\toprule
 & \multicolumn{2}{c}{OHTS} & \multicolumn{2}{c}{GRAPE (PLR3)} & \\ \cmidrule(rl){2-3}\cmidrule(rl){4-5}
& ACC & AUC & ACC & AUC & PARAMs \\ \midrule
input size & 0.888 & 0.927 & \textbf{0.947} & 0.955 & \textbf{2,956,036} \\
2 * input size  & \textbf{0.891} & \textbf{0.935} & \textbf{0.947} & \textbf{0.956} & 4,136,452 \\
self-attention & 0.887 & 0.903 & \textbf{0.947} & 0.954 & 8,855,044 \\ \bottomrule
\end{tabular}

\label{tab:ablation2_ohts_grape}
\end{table}

\section{Conclusion}
In this paper, we propose TSDF, a two-stage framework for variable-length glaucoma prognosis. Specifically, we decouple this task into two stages: The first stage employs a self-supervised masked autoencoder for feature representation, enabling joint training across multiple glaucoma datasets and improving feature learning for small datasets. The second stage introduces a temporal aggregation module with masked non-local attention layers, efficiently aggregating variable-length sequences. Its dual-path design captures both single-frame and time-series frame information. We validate our approach on two glaucoma datasets of different scales, OHTS and GRAPE, demonstrating superior performance.

\vspace{1em}
\textbf{Acknowledgement.}
This work was supported in part by the National Eye Institute under grant R21EY035296, and in part by the National Science Foundation under grant 2306556. The content is solely the responsibility of the authors and does not represent the official views of the National Institutes of Health.

\bibliographystyle{splncs04}
\bibliography{mybibliography}
\typeout{get arXiv to do 4 passes: Label(s) may have changed. Rerun}
\end{document}